\documentclass[conference]{IEEEtran}
\IEEEoverridecommandlockouts
\usepackage{cite}
\usepackage{amsmath,amssymb,amsfonts}
\usepackage{algorithmic}
\usepackage{graphicx}
\usepackage{textcomp}
\usepackage{xcolor}
\usepackage{makecell}
\usepackage{subcaption}
\usepackage{indentfirst}
\usepackage{hyperref}
\hypersetup{hidelinks,
	colorlinks=true,
	citecolor = green,
	urlcolor = magenta,
	pdfstartview=Fit,
	breaklinks=true}

\def\BibTeX{{\rm B\kern-.05em{\sc i\kern-.025em b}\kern-.08em
    T\kern-.1667em\lower.7ex\hbox{E}\kern-.125em X}}
\begin{document}

\title{High Dynamic Range Imaging with
Context-aware Transformer\\
\thanks{* These authors contributed to the work equallly and should be regarded as co-first authors.}
}

\author{\IEEEauthorblockN{1\textsuperscript{st} Fangfang Zhou*}
\IEEEauthorblockA{
	\textit{Senslab Technology Co., Ltd.}\\
	Shanghai, China \\
	51174700059@stu.ecnu.edu.cn}
\and
\IEEEauthorblockN{2\textsuperscript{nd} Zhengming Fu}
\IEEEauthorblockA{
	\textit{NeuroSens Technology}\\
	Austin, U.S. \\
	zenith.fu@neurosens.ai}
\and
\IEEEauthorblockN{3\textsuperscript{rd} Dan Zhang*}
\IEEEauthorblockA{
	\textit{Senslab Technology Co., Ltd.}\\
	Shanghai, China \\
	zhangdan\underline{ }fiona@163.com}
}


\maketitle

\begin{abstract}
Avoiding the introduction of ghosts when synthesising LDR images as high dynamic range (HDR) images is a challenging task. Convolutional neural networks (CNNs) are effective for HDR ghost removal in general, but are challenging to deal with the LDR images if there are large movements or oversaturation/undersaturation. Existing dual-branch methods combining CNN and Transformer omit part of the information from non-reference images, while the features extracted by the CNN-based branch are bound to the kernel size with small receptive field, which are detrimental to the deblurring and the recovery of oversaturated/undersaturated regions. In this paper, we propose a novel hierarchical dual Transformer method for ghost-free HDR (HDT-HDR) images generation, which extracts global features and local features simultaneously. First, we use a CNN-based head with spatial attention mechanisms to extract features from all the LDR images. Second, the LDR features are delivered to the Hierarchical Dual Transformer (HDT). In each Dual Transformer (DT), the global features are extracted by the window-based Transformer, while the local details are extracted using the channel attention mechanism with deformable CNNs. Finally, the ghost free HDR image is obtained by dimensional mapping on the HDT output. Abundant experiments demonstrate that our HDT-HDR achieves the state-of-the-art performance among existing HDR ghost removal methods.
\end{abstract}

\begin{IEEEkeywords}
HDR deghosting, Transformer, CNN, Attention
\end{IEEEkeywords}

\section{Introduction}
HDR imaging methods are to produce an image with a wide dynamic range, which is closer to the human eye's perception and generated by multiple low dynamic range (LDR) images with varying exposures. If the LDR images are perfectly aligned, i.e. no camera shaking or object moving, we can fuse the LDR images directly and obtain perfect HDR images without ghost. However, this condition can be extremely difficult to achieve in the real-world.
\begin{figure}[htbp]
	\setlength{\abovecaptionskip}{0cm}
	\begin{subfigure}{0.495\linewidth}
		\centering
		\includegraphics[height=1.389in]{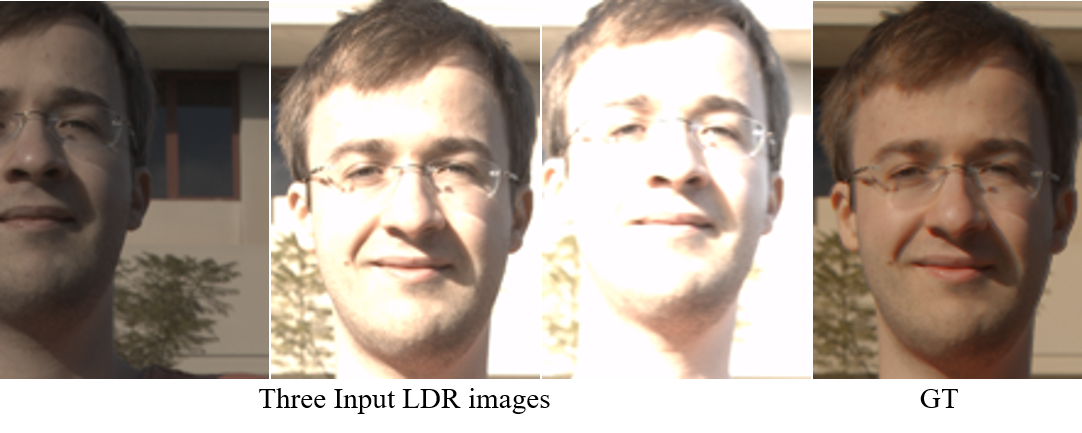}
	\end{subfigure}
	
	\begin{subfigure}{0.235\linewidth}
		\centering
		\includegraphics[height=1.215in]{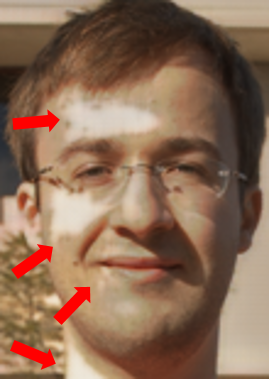}
		\setlength{\abovecaptionskip}{-0.35cm}
		\captionsetup{font=scriptsize}
		\caption*{DeepHDR \cite{b13}}
	\end{subfigure}
	\begin{subfigure}{0.235\linewidth}
		\centering
		\hspace{-0.04cm}
		\includegraphics[height=1.215in]{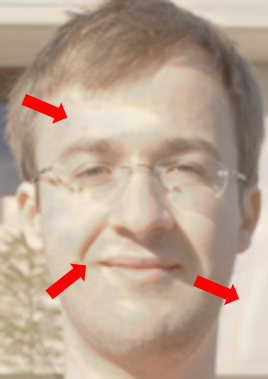}
		\setlength{\abovecaptionskip}{-0.35cm}
		\captionsetup{font=scriptsize}
		\caption*{AHDRNet \cite{b17}}
	\end{subfigure}
	\begin{subfigure}{0.235\linewidth}
		\centering
		\hspace{-0.04cm}
		\includegraphics[height=1.215in]{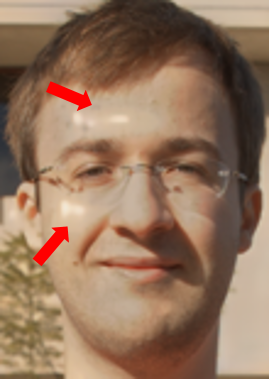}
		\setlength{\abovecaptionskip}{-0.35cm}
		\captionsetup{font=scriptsize}
		\caption*{HDR-GAN \cite{b18}}
	\end{subfigure}
	\begin{subfigure}{0.24\linewidth}
		\centering
		\hspace{-0.04cm}
		\includegraphics[height=1.215in]{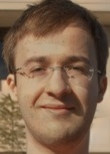}
		\setlength{\abovecaptionskip}{-0.35cm}
		\captionsetup{font=scriptsize}
		\caption*{Ours}
	\end{subfigure}
	\setlength{\belowcaptionskip}{-15pt}
	\caption{Qualitative comparison of three CNN-based methods with ours on the dataset of Kalantari et al. \cite{b12}. As shown above, our proposed HDT-HDR, which is used to extract global features and local features simultaneously, produces better results that are free from ghost artifacts and recover more details in the oversaturated regions.}
	\label{fig1}
\end{figure}

Most of traditional HDR algorithms need to discard the unaligned pixels \cite{b1, b2, b3, b4} or align all pixels \cite{b5, b6, b7} in LDR images before fusing them. The former is to register all pixels in the LDR images, mark the unaligned parts, and remove the unaligned areas or replace them with pixels in the reference image. The HDR image generated by these methods will lose a lot of information in areas where pixels are shifted. For the latter methods, the key to producing a high-quality HDR image is to find an appropriate way to align other LDR images with the reference image. However, traditional image alignment methods, such as optical flow methods \cite{b8}, patch-based optimization methods \cite{b10, b11} and mesh flow methods \cite{b9}, will inevitably produce ghosts because they cannot strictly align when large movements occur.

In recent years, CNN-based algorithms \cite{b31,b32,b33,b34,b17,b21,b41,b45, b49} have been proved to be significantly superior to the traditional algorithms \cite{b42,b43,b44}, in terms of both performance and computational cost. Liu et al. \cite{b46} proposed an improved YOLOv5 network architecture for agriculture image recognition. Especially, a tiny detector layer was introduced to enhance the performance. Bian et al. \cite{b47} and Song et al. \cite{b48} use CNN-based methods for medical images reconstruction. Kalantari et al. \cite{b12} were the first to propose using deep learning to generate HDR images. This method required first aligning LDR images using optical flow, and then synthesising HDR images with the aligned LDR images using a CNN-based method. Wu et al. \cite{b13} used the homography transformation alignment method in the early stage of HDR imaging, followed by a CNN-based method. Like traditional algorithms, such two-stage HDR methods would introduce ghosting due to the inability to strictly align when there are large moving, oversaturated, or occluded areas exist in LDR images. To solve this problem, many researchers \cite{b14, b15, b16, b17, b18} no longer align LDR images before inputting them into the networks, but take the unaligned LDRs and their gamma-corrected images directly as inputs to produce the HDR image. These methods utilize massive data-driven models to implicitly learn to align LDR images and synthesise the corresponding HDR image. This type of end-to-end HDR deghosting algorithm can achieve high-quality deghosting in most cases. However, when there are big movements or oversaturation in LDR images, HDR imaging will be accompanied by ghost or motion blur, as shown in Fig.~\ref{fig1}. This is because the intrinsic properties of CNNs, such as parameter sharing and small receptive fields, determine their weak ability to deal with problems sensitive to global information \cite{b19, b20}. Therefore, to generate high-quality ghost-free HDR images when LDR images have large movements or oversaturation, CNN-based methods alone will not work.

Vision Transformer (ViT) \cite{b40} is known for its long-range modelling capability, and can flexibly adjust its receptive field. However, compared to CNN-based models, ViT requires a larger amount of data to train due to its lack of bias and weaker ability to extract local context. Therefore, some scholars \cite{b22, b39} have tried to combine Transformer with CNN to get better performance. Zhen Liu et al. \cite{b22} proposed a method called HDR-Transformer, which combines CNN and ViT (CA-ViT) for HDR deghosting. They divided the model into two branches, one branch using Transformer to extract global information, and the other branch using CNN-based channel attention mechanisms to supplement local information. However, HDR-Transformer did not apply spatial attention to the reference image during the shallow feature extraction, thus losing the opportunity for initial recovery of the oversaturated or undersaturated areas in the reference image. And the CA-ViT channel attention mechanism uses ordinary convolution, which can only extract features within a fixed kernel size, which is not conducive to recovering blur caused by small movements.

Based on these situations, we propose a novel HDR deghosting method, which mainly consists of hierarchical dual branches built with deformable CNNs and transformers. We take the medium exposure image as the reference image. Firstly, we use the spatial attention mechanism to extract the shallow features of all the LDR images, including the reference image, and concatenate them. It is worth noting that we use the spatial attention of the features of the long and short exposure images on the reference image. This is advantageous to make full use of the information in the long and short exposures to correct the oversaturated or undersaturated areas in the mid-exposure image.  Secondly, the concatenated features are supplied into the HDT, where deformable convolutions can help the local branch to obtain different features with different receptive field to avoid the blur caused by small movements, while the Transformer in the global branch captures the global information, such as the long range movements. Finally, the local and global information are fused and passed through a convolution layer for channel transformation, and the final ghost-free HDR image is obtained. Our main contribution can be summarised as follows:

1. We show that adding spatial attention to the reference image can partially compensate for missing information caused by oversaturation or undersaturation.

2. We propose a novel hierarchical dual Transformer, named HDT-HDR, which uses deformable CNNs to extract the local texture information for small motion deblurring and Transformer to capture the global information for large motion deghosting.

3. Abundant experiments prove that our HDT-HDR outperforms the state-of-the-art in HDR deghosting algorithms both qualitatively and quantitatively.

\section{Related works}

To produce high-quality ghost-free HDR images, a large number of methods \cite{b2, b12, b14, b26, b29, b30, b31, b32} have been proposed by scholars.  These methods can be classified into three broad categories: Traditional methods \cite{b1, b23, b24, b27, b28}, CNN-based methods \cite{b12, b13, b32, b33, b34} and Transformer-CNN-based method \cite{b22}.

{\bf Traditional methods} Traditional methods are generally divided into two classes: motion rejection and motion registration. Under the assumption that most of the pixels in the LDR images are static and only a small number of pixels move, the first type is to register all the pixels in the LDR images, select the unaligned areas, replace them with pixels in the reference image or surrounding static pixels, and then fuse the aligned LDR images to generate HDR images. Grosch et al. calculated the predicted pixels according to the brightness consistency criterion and compared them with the real pixels to generate an error map to identify moving objects \cite{b1}. Gallo et al. used the consistency of patches in all exposures with the reference image to distinguish the moving pixels \cite{b2}. Jacobs et al. \cite{b23} and Reinhard et al. \cite{b24} used weighted irradiance map variance and intensity variance, respectively. Khan et al. calculated the probability maps \cite{b25}. Oh et al. \cite{b26} detected ghost regions by rank minimization. The second type focuses on the methods of aligning LDR images with the corresponding reference image, and then fusing the aligned LDR images. Bogoni et al. \cite{b27} predicted the motion vectors in multi-scale by optical flow to align LDR images. Kang et al. \cite{b28} mapped LDR image intensities to the luminance domain and used optical flow to align them. Jinno and Okuda \cite{b29} aligned the input images with dense corresponding maps predicted by Markov Random Field. Hu et al. \cite{b30} calculated the brightness and gradient consistencies to align input images in the transformed domain. For the first type of method, HDR imaging does not perform well due to the masking of a large amount of information. For the second type of methods, the resulting HDR image would be accompanied by ghosting or motion blur, when objects have a large moving or saturation in LDR images.

\begin{figure*}[htbp]
	\setlength{\abovecaptionskip}{0cm}
	\begin{subfigure}{0.99\linewidth}
		\centering
		\includegraphics[height=2.185in]{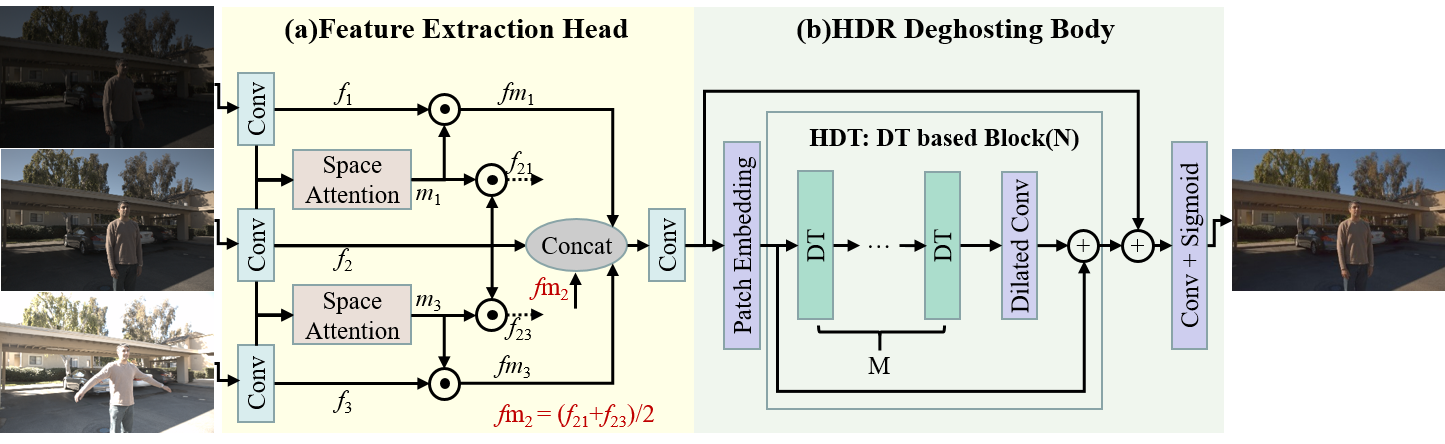}
	\end{subfigure}
	\setlength{\belowcaptionskip}{-10pt}
	\caption{The network architecture of HDT-HDR. The pipeline composed of two stages: (a) The feature extraction head, it uses the  spatial attention module to extract the coarse features. (b) The HDR deghosting body, which consists of several DT based Blocks, the extracted coarse features are fed into it to recover the HDR images.}
	\label{fig2}
\end{figure*}

{\bf CNN-based methods} In recent years, a large amount of CNN-based methods have been put forward. Eilertsen et al. \cite{b31} adopted U-Net to map a single LDR image directly to an HDR image. Lee et al. \cite{b32} utilized GAN to generate pseudo multi-exposed images from a single image, and HDR image could be pixel-wisely fused by them. Kalantari et al. \cite{b12} first introduced a CNN-based method in HDR imaging after aligning multi-exposed LDR images with optical flow. Wu et al. \cite{b13} first used a simple homography transformation to align the background and generated an HDR image from the aligned multi-exposed LDR images with U-Net based networks. Pu et al. \cite{b33} applied deformable convolutions for pyramidal alignment, squeeze-and-excitation (SE) attentionto to fully exploit aligned features and mask-based weighting for refining HDR image reconstruction. Prabhakar et al. \cite{b14} proposed an efficient method with bilateral-guiding upsampler to generate HDR images. Yan et al. \cite{b17} adopted a spatial attention to reduce the ghost on HDR image and Yan et al. \cite{b34} used non-local blocks to capture the global information of unaligned inputs to help LDR images better aligned. Niu et al. \cite{b18} first employed a GAN-based method to synthesize HDR images by fusing multi-exposed LDR images. Liu et al. \cite{b16} applied a spatial attention module to handle multi-saturation, and a Pyramid, Cascading and Deformable (PCD) alignment module to tackle misalignments. Chung et al. \cite{b15} transformed the problem of motion alignment to the brightness adjustment to align images for next fusing. All the above methods, using only a single LDR image, will inevitably produce low-quality HDR image due to the lack of real exposure information. What’s more, the inherent properties of CNN-based methods lack global information, and would not solve the ghosting problem for HDR imaging well when large objects are moving or extremely oversaturated/undersaturated regions exist in LDR images.

{\bf Transformer-CNN-based methods} Transformers have been hugely successful in the field of natural language processing \cite{b20}. After embedding words, the attention mechanism of multiple heads is used to obtain a long range of connections for processing natural language. Recently, ViT \cite{b40} proved that the pure Transformer can achieve a comparable result to CNN networks in image classification tasks by treating image patches as words and adding tokens to denote the category. With the development of Swin-Transformer \cite{b35}, which used the shift-window scheme to greatly reduce computational cost and made full use of image information, many Transformer-based algorithms emerged in computer vision. Liang et al. \cite{b19} proposed SwinIR for image super-resolution and denoising, and achieved the state-of-the-art performance. Liu et al. \cite{b22} not only used CNN, but also added Transformer in their HDR-Transformer architecture, combining the advantages of CNN in extracting local information and Transformer in capturing global information. However, HDR-Transformer did not do spatial attention on the reference image during the shallow feature extraction, which is not beneficial for recovering the oversaturation/undersaturation region, and the CNN-based branch has limited receptive field and is not good at local texture changes caused by small movements. Inspired by \cite{b22, b33}, we propose HDT-HDR based on Transformers and deformable CNNs.

\section{Method}
Our goal is to use LDR images with multiple exposure times to produce the corresponding ghost-free HDR images. Following the previous studies \cite{b12, b13, b17}, three LDR images with different exposure times ($I_{i}$, i = 1, 2, 3) are used and the intermediate image $I_{2}$ is used as the reference image in this paper.

Firstly, in order to make full use of the image information, the input images are mapped to the HDR space to obtain the corresponding gamma-corrected images $\tilde{I}_{i}$.
\begin{equation}
	\tilde{I_{i}} = \frac{(I_{i})^{\gamma}}{t_{i}}, i=1,2,3
     \label{eq1}
\end{equation}
Where  $\gamma$ denotes the gamma correction parameter ($\gamma=2.2$ in this paper), and $t_{i}$ denotes the exposure time of $I_{i}$. According to \cite{b12}, we simply concatenate $I_{i}$ and $\tilde{I}_{i}$ on the channel dimension, forming the input $I_{ci}$ with 6 channels of the network. This method of feature enhancement not only can employ LDR images to reduce noise and identify saturation areas, but also can use gamma-corrected images to help with image alignment. 

Secondly, the HDR image $I_{H}$ are generated with our HDR deghosting model $F(.)$:
\begin{equation}
	{I_{H}} = F(I_{ci};\theta), i=1,2,3
    \label{eq2}
\end{equation}
Where $\theta$ represents the model parameters.

Finally, according to \cite{b11}, we usually need to display the tonemapped HDR images in practical applications. In order to generate better HDR images, it is better to optimize the model with the loss calculated in the tonemapping domain through tonemapping the model output $I_{H}$ and the label image $I_{GT}$ by a certain rule. This paper uses $\mu$-law for tonemapping:
\begin{equation}
	T(x) = \frac{log(1+\mu x)}{log(1+\mu)}
    \label{eq3}
\end{equation}
$\mu$ determines the compression degree ($\mu$=5000, in this paper). Our model is optimized by $L_{1}$  loss:
\begin{equation}
	L = \parallel{T(I_{GT})-T(I_{H})}\parallel_{1}
    \label{eq4}
\end{equation}

The entire network architecture is shown in Fig.~\ref{fig2}. HDT-HDR is mainly consists of Feature-Extraction head and HDR-Deghosting body. In the Feature-Extraction head, we first extract features by convolutional layers, and then use the spatial attention mechanism to reduce the interference of moving objects and initially correct the saturation area of the reference image. It is worth noting that this paper exploits the spatial attention of non-reference images and reference image, and performs attention filtering on the features of the reference image to better utilize the information of non-reference images. The HDR deghosting bodies are realized by hierarchical dual-branch Transformers, in each of which, the local features and global information are extracted and fused simultaneously. Among them, the global branch can extract the global information to align the input images, while the local branch uses the convolutional layer to extract local features and the channel attention mechanism with deformable convolution to filter the importance of local textures. Then, the fusion of the features of the two branches  can not only use  the global information to ensure that the HDR image does not contain ghosts which caused by object long-distance movement in LDR images, but also use the local information to avoid small motion blur in the generated HDR image.

\begin{figure}[t]
	\setlength{\abovecaptionskip}{0cm}
	\centering
	\includegraphics[width=0.7\linewidth]{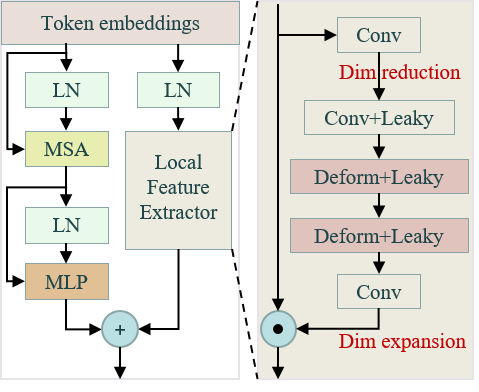}
	\setlength{\belowcaptionskip}{-10pt}
	\caption{Illustration of the proposed dual-branch Transformer architecture DT. Multi-head Transformer encoder is used to obtain the context of all input images and extract the global information. The channel attention mechanism with deformable convolution is used to extract the local feature information of all images and the image connection between frames. In the end, the features from two branches are fused as DT output.}
	\label{fig3}
\end{figure}

The detail of our proposed DT structure is introduced in Fig.~\ref{fig3}.

\subsection{Feature-Extraction head}
First, preliminary feature extraction and channel transformation on the input $I_{ci}\in R^{H\times W\times 6} (i=1, 2, 3)$ to obtain $f_{i} \in R^{H\times W\times C}$ by three convolutional layers, $C$ denotes the number of channels. Second, two spatial attention modules are used to calculate the attention map of non-reference image features $f_{i} (i=1,3)$ and the reference image feature $f_{2}$, respectively. Third, all non-reference images $f_{i} (i=1, 3)$ are multiplied by the corresponding attention map $m_{i} (i =1, 3)$, so as to initially align the non-reference images with the reference image. The alignment process can be summarized as the follows:
\begin{equation}
	\setlength{\abovedisplayskip}{0pt}
	\setlength{\belowdisplayskip}{5pt}
	m_{i} = Att(f_{i}, f_{2}), i=1,3
	\label{eq5}
\end{equation}
\begin{equation}
	fm_{i} = f_{i}\odot m_{i}, i=1,3
	\label{eq6}
\end{equation}

$\odot$ denotes element-wise multiplication, $fm_{i}$ denotes the features output of the spatial attention. Unlike \cite{b22}, we also calculate the average feature of spatial attention on $f_{2}$ by $m_{1}$ and $m_{3}$, thereby the missing information in $f_{2}$ due to the saturated region in the reference image can be supplied, and the output is denoted as $fm_{2}$. We summarize the process as \eqref{eq6}.
\begin{equation}
	fm_{2} = (f_{2}\odot m_{1} + f_{2}\odot m_{3})/2
	\label{eq6}
\end{equation}

Then, all $fm_{i} (i=1, 2, 3)$ as well as $f_2$ are concatenated on the channels to obtain $f_{init}$ for the next process.

\subsection{HDR-Deghosting body}
As in \cite{b22}, our main module is Hierachical Dual-branch Transformer (HDT), and each DT is a parallel two-branch network structure composed of the Transformer-based global branch and the CNN-based local branch. The HDR-Deghosting body embeds $f_{init}$ and takes the embeddings $Em_{0}$ into HDT. After several skip-connections and convolution layers we will obtain the final output, as shown in Fig.~\ref{fig2}.

{\bf Transformer-based Global Branch} Following \cite{b22}, we obtain the long-range information through a window-based multi-head Transformer encoder, which is composed of several LNs, a multi-head self-attention (MSA) module, a multi-layer perceptron (MLP) and several residual connections.
Embedded feature $Em_{0}$ will be taken as the input, and global feature extraction can be described as:
\begin{equation}
	Em_{1} = MSA(LN(Em_{0})) + Em_{0}
\end{equation}
\begin{equation}
	GF_{global} = MLP(LN(Em_{1})) + Em_{1}
\end{equation}
Where, $GF_{global}$ represents the global information.

\begin{figure*}[htbp]
	\setlength{\abovecaptionskip}{0cm}
	\begin{subfigure}{0.99\linewidth}
		\centering
		\includegraphics[width=7.12in]{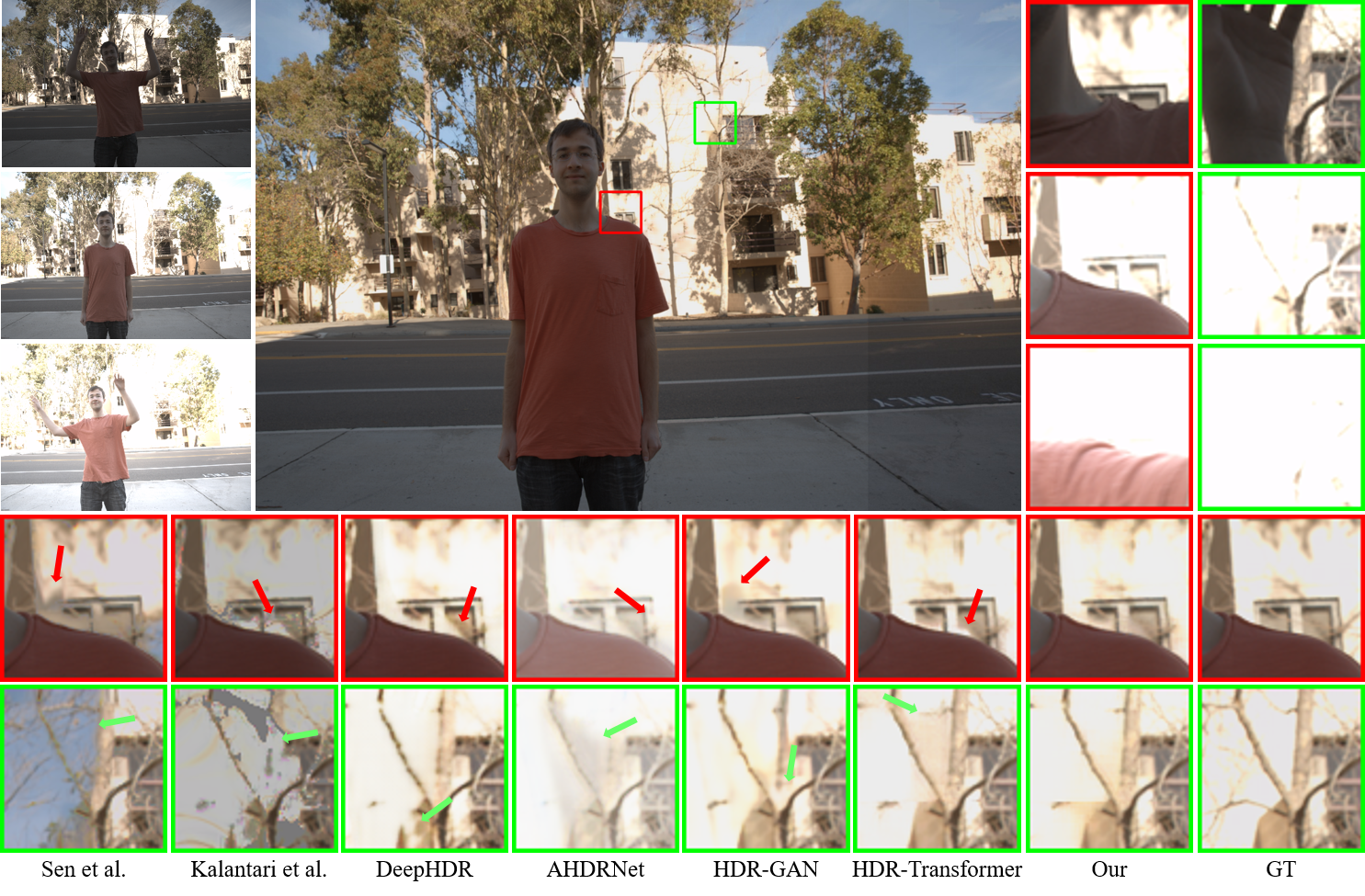}
	\end{subfigure}
	\setlength{\belowcaptionskip}{-15pt}
	\caption{Visual comparison of the state-of-the-art methods \cite{b36,b12, b13, b17, b18, b22} on Kalantari et al \cite{b12}’s test set. As shown, the traditional methods \cite{b36} have obvious disadvantages, while the CNN-based methods could not remove the long-range ghosts \cite{b12, b17} or generate local details in oversaturated regions \cite{b13, b18}. Transformer and CNN combined method \cite{b22} performs better than CNN-based methods but still faces the same problems as the CNN-based methods. Among them, our proposed HDT-HDR has the best performance both in texture information recovery and deghosting.}
	\label{fig4}
\end{figure*}

{\bf CNN-based Local Branch} We extracted the local information by a CNN-based module using the channel attention mechanism. For the token embedding vector $Em_{0}$, we normalize it by LN layer, and reshape it to a vector shaped in $N\times H\times W\times C$. And we realize the vector dimensionality reduction by an ordinary convolution and get a feature vector shaped in $N\times H\times W\times C/10$. Then we use another ordinary convolution to undertake the feature. Each convolution except the one after the LN layer will be followed by a LeakyReLU activation layer to better select features. Next, we use two convolutions with deformable kernels \cite{b38} to fuse the context details with changes in surroundings and the cross-channel feature, which is a key design to avoid small motion blur. Now we have a vector whose shape is $N\times H\times W\times 2C/5$. We do an average pooling on it and get a vector with a shape of $N\times 2C/5$. We perform linearzation and sigmoid on it to obtain a vector with shape of $N\times C$, which can be treated as the attention channel weight $w_{c}$, and will be multiplied with the $f_{in}$, which is a processed $Em_{0}$ to accommodate subsequent convolution operations. The final local output is now obtained. The local feature extraction process can be described as follows:
\begin{equation}
	f_{in} = Reshape(LN(Em_{0}))
\end{equation}
\begin{equation}
	f_{local} = \sigma_{1}(D(\sigma_{1}(D(\sigma_{1}(Conv(\sigma_{1}(Conv(f_{in}))))))))
\end{equation}
\begin{equation}
	w_{c} =\sigma_{2} (FC(f_{local}))
\end{equation}
\begin{equation}
	LF_{local} = w_{c} \odot f_{in}
\end{equation}
$\sigma_{1}$ denotes LeakyReLu, $\sigma_{2}$ denotes sigmoid, $D$ denotes deformable convolutional layer and $LF_{local}$ represents the local information extracted by this branch.

The global feature information and the local feature information are fused by element-wise addition to $f_{fusion}$, and the forward propagation of one DT module is completed. Then $f_{fusion}$ would be the input for the next DT. DT repeats N times (N = 6 in this paper), forming DT groups, and M DT groups (M = 3 in this paper) form the main body of HDT. HDT output passes through a dilated convolutional layer which can expand the receptive field, participate in two global residuals in succession, pass through several convolutional layers and a sigmoid layer, and finally the deghosting HDR image can be obtained. The overall architecture is shown in Fig.~\ref{fig2}.

\section{Experiments}
\subsection{Datasets, Metrics and Implementation Details}
{\bf Datasets} Following \cite{b13, b17,b18,b22}, we use Kalantari et al.'s dataset \cite{b12} as the training, validation and test sets for the experiments. Kalantari et al.'s dataset contains 74 training samples and 15 test samples, and each sample contains three LDR images with different exposure times as well as the corresponding ground truth HDR image. We also use Sen et al. \cite{b36}'s and Tursun et al. \cite{b37}'s datasets, which do not contain ground truth images, as test sets for visually evaluating the generalization ability of our model.

Before training, we apply the horizontal sliding window with a step size of 64, crop the original image into small patches with size of 128×128, and then use rotation and flipping for data enhancement.

{\bf Metrics} We use PSNR and SSIM as evaluation metrics for quantitative comparison. We calculate ${\rm PSNR}_{\mu}$ and ${\rm SSIM}_{\mu}$ between the model output image $I_{H}$ and the ground truth image $I_{GT}$ in their tone-mapped domain, as well as ${\rm PSNR}_{l}$ and ${\rm SSIM}_{l}$ in the linear domain. To evaluate the visibility and quality of the synthesized HDR images under different luminance conditions, we also calculate HDR-VDP-2 according to \cite{b18}.

{\bf Implementation Details} All of our experiments are conducted on pytorch 3.8, an NVIDIA Tesla T4 GPU with 16GB per GPU. Our HDR deghosting model training adopts ADAM optimizer. The initial learning rate is 1e-4, $\beta 1 = 0.9, \beta 2 = 0.999$ and $\epsilon = 10^{-8}$. The maximum number of epochs is 100, and the batch size is 16. Our model takes about 46 hours to train.

\subsection{Comparison with State-of-the-art Methods}
{\bf All Compared models} To evaluate our model, we compared it with several state-of-the-art HDR deghosting methods, including two traditional HDR algorithms as Sen et al. \cite{b36} and Hu et al. \cite{b11}, CNN-based algorithms as Kalantari et al. \cite{b12}, DeepHDR \cite{b13}, AHDRNet \cite{b17}, NHDRRNet \cite{b14},  and HDR-GAN \cite{b18}, a single Transformer algorithm SwinIR \cite{b19} and a combined CNN-Transformer algorithm HDR-Transformer \cite{b22}.

\begin{table}[t]
	\caption{Quantitative comparison with state-of-the-art methods on Kalantari et al. \cite{b12}’s test samples. PSNR, SSIM, and HDR-VDP-2 are as the metrics used to evaluate the models. '$\mu$' and '$l$' represent the values calculated on the tonemapped domain and the linear domain, respectively.  The best results are {\bf black bold}, while the second is \underline {underlined}.}
	\setlength{\abovecaptionskip}{0cm}
	\setlength{\belowcaptionskip}{0cm}
	\centering
	\begin{tabular*}{\hsize}{l@{\extracolsep{\fill}}l@{\extracolsep{\fill}}l@{\extracolsep{\fill}}c@{\extracolsep{\fill}}c@{\extracolsep{\fill}}c@{\extracolsep{\fill}}}
		\hline
		Methods & ${\rm PSNR}_{\mu}$ & ${\rm PSNR}_{l}$ & ${\rm SSIM}_{\mu}$ & ${\rm SSIM}_{l}$ & HDR-VDP-2 \\
		\hline
		Sen et al.\cite{b36}	&40.80 &38.11	&0.9808	&0.9721	&59.38 \\
		Hu et al.\cite{b11}	&35.79	&30.76	&0.9717	&0.9503	&57.05\\
		Kalantari et al. \cite{b12} &42.67	&41.23	&0.9888	&0.9846	&65.05\\
		DeepHDR\cite{b13}	&41.65	&40.88	&0.9860	&0.9858	&64.90 \\
		AHDRNet\cite{b17}	&43.63	&41.14	&0.9900	&0.9702	&64.61 \\
		NHDRRNet \cite{b34}	&42.41	&-	&0.9887	&-	&61.21 \\
		HDR-GAN\cite{b18}	&43.92	&41.57	&0.9905	&0.9865	&65.45\\
		SwinIR\cite{b19}	&43.42	&41.68	&0.9882	&0.9861	&64.52 \\
		HDR-Transformer\cite{b22}	&\underline {44.21} 	&\underline{42.17}	&\underline{0.9918}	&\underline{0.9889}	&\underline{66.03} \\
		Ours	&{\bf 44.36}&{\bf 42.73}&{\bf 0.9919}&{\bf 0.9898} & {\bf 66.08}\\
		\hline
	\end{tabular*}
	
	\label{table 1}
\end{table}
				
{\bf Quantitative comparison} To quantitatively compare our model with other state-of-the-art methods, we present the average results for each model on the 15 test samples from the Kalantari et al.'s dataset in Table \ref{table 1}, and all values but ours are from HDR-Transformer. \cite{b22}. Several conclusions can be drawn from Table \ref{table 1}. First, traditional models have obvious disadvantages. Second, the pure Transformer-based model can compete with most of CNN-based models to some extent, but is inferior to HDR-GAN. Third, HDR-Transformer that combined with Transformer and CNN performs better than using CNN or Transformer alone. Finally, our HDT-HDR performs better than HDR-Transformer \cite{b22}.

{\bf Qualitative comparison} To further verify the visual performance of our model, qualitative comparisons are made with test samples from Kalantari et al.'s, Sen et al.'s and Tursun et al.'s datasets. The images are generated using the author's models when the pre-trained model is available online. Otherwise we retrain the model ourselves using the official realization. Fig.~\ref{fig4} shows the performance of several models on synthetic HDR image with long range of object motion and oversaturation in Kalantari et al.’s test samples. The first row in the figure shows the LDR images, the HDR image generated by our model, and the zoomed LDR patches that contain large range object movement and oversaturation, respectively. The second and third rows show the zoomed HDR patches generated by different models. In the second row with red borders, that other models are not so successful in processing the oversaturated area at the red arrow, while the texture information generated by our model is closer to the GT image. In the third row with green borders, the images generated by Sen et al., Kalantari et al. and AHDRNet have severe ghosting, while DeepHDR and HDR-GAN have much texture information missing in the oversaturated area. The remaining models also have slight ghosting and texture loss in visual. Generally, our model performs better on the Kalantari et al.'s datasets, both in terms of deblurring and deghosting. 

\begin{figure*}[htbp]
	\setlength{\abovecaptionskip}{0cm}
	\begin{subfigure}{0.495\linewidth}
		\centering
		\includegraphics[width=3.55in]{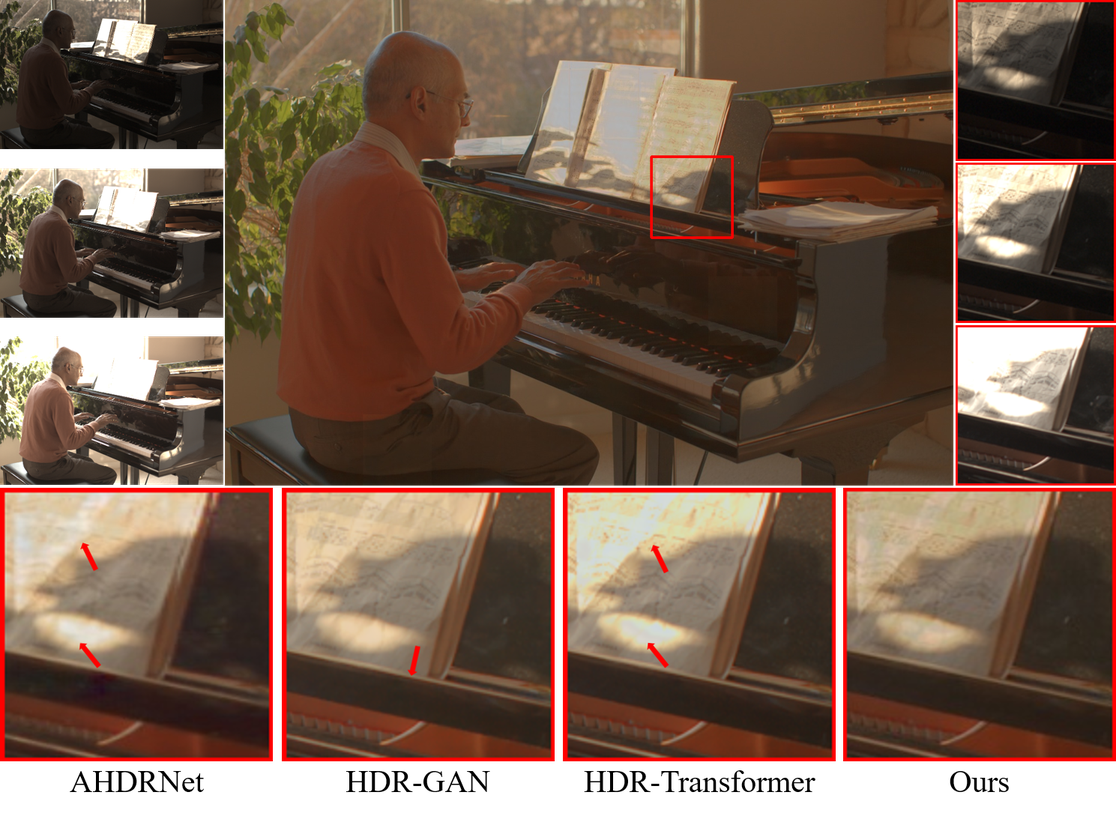}
		\setlength{\abovecaptionskip}{-0.5cm}
		\caption{Sen et al.'s dataset \cite{b36}}
		\label{fig5a}
	\end{subfigure}
	\begin{subfigure}{0.495\linewidth}
		\centering
		\includegraphics[width=3.55in]{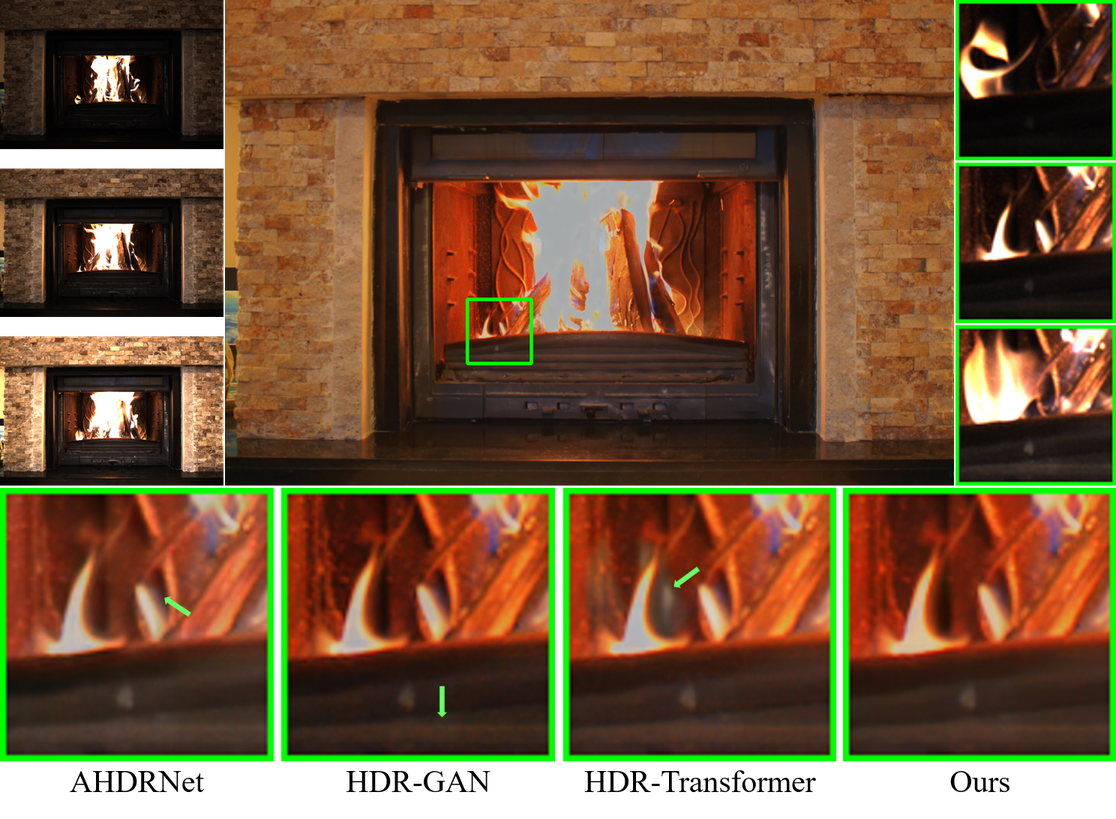}
		\setlength{\abovecaptionskip}{-0.5cm}
		\caption{Tursun et al.'s dataset \cite{b37}}
		\label{fig5b}
	\end{subfigure}
	\setlength{\belowcaptionskip}{-15pt}
	\caption{Visual comparison of the state-of-the-art methods \cite{b36,b12, b13, b17, b18, b22} on  Sen et al. \cite{b36}'s and Tursun et al. \cite{b37}'s datasets. When there are large areas of oversaturation and undersaturation in LDR images, our model generates HDR images with better detail restoration and boundary preservation than other models.}
	\label{fig5}
\end{figure*}

Fig.~\ref{fig5} shows the visual comparison of HDR images composited by several models from Sen et al.’s and Tursun et al.’s datasets. Fig.~\ref{fig5a} shows the HDR images generated by different methods when the LDR images contain large oversaturated regions. The second row of Fig.~\ref{fig5a} shows that small motion deblurring of the HDR-GAN model on the oversaturated region is slightly better than our model. But from the area pointed by the red arrow, it can be seen that the HDR-GAN performs poorly in undersaturated region. Fig.~\ref{fig5b} shows the HDR images generated by the same models when the LDR images contain large areas of undersaturated regions. As can be seen from the green arrow, the image generated by AHDRNet contains slight ghosting, the image generated by HDR-GAN still contains burr, and the image generated by HDR-Transformer contains obvious large-area ghosting, while our method performs well both in oversaturated region and undersaturated regions. 
\begin{table}[b]
	\setlength{\abovecaptionskip}{0cm}
	\setlength{\belowcaptionskip}{0cm}
	\centering
	\caption{The inference times and parameters of different models. Part of the values are from \cite{b22}.}
	\begin{tabular*}{\hsize}{l@{\extracolsep{\fill}}l@{\extracolsep{\fill}}c@{\extracolsep{\fill}}c@{\extracolsep{\fill}}}
		\hline
		Methods & Environment & Time(s)	&Parameters(M) \\
		\hline
		Sen et al. \cite{b36}	&CPU	&61.81	&-\\
		Hu et al. \cite{b11}	&CPU	&79.77	&-\\
		Kalantari et al. \cite{b12}	&CPU+GPU	&29.14	&0.3\\
		DeepHDR \cite{b13}	&GPU	&0.24	&20.4\\
		AHDRNet \cite{b17}	&GPU	&0.30	&1.24\\
		NHDRRNet \cite{b34}	&GPU	&0.31	&38.1\\
		HDR-GAN \cite{b18}	&GPU	&0.29	&2.56\\
		HDR-Transformer \cite{b22}	&GPU	&0.15	&1.22\\
		Ours 	&GPU & 0.16 & 1.35\\
		\hline
	\end{tabular*}
	\label{table 2}
\end{table}

{\bf Analysis of Computational Budgets} We compare the amount of parameters of each model and the test time on same data set in Table \ref{table 2}. It can be seen from the table that the traditional HDR algorithms take even more than one minute, which is unbearable in practical applications. The inference time of the CNN-based methods is significantly reduced, and our model size is even smaller. Our model can inference faster than the CNN-based models. Although our model size is slightly larger than the HDR-Transformer, it can be ignored considering of its super HDR imaging performance.

\subsection{Ablation Study}
All our ablation experiments are conducted on Kalantari et al.’s\cite{b12} dataset, PSNR and HDR-VDP-2 are used as quantitative evaluation metrics.

{\bf Ablation on the network architecture} For the network design, we compare the proposed DT, the spatial attention on reference images (SAR) module, and the entire HDT-HDR with the baseline. In detail, we design the following variants:

\begin{table}[t]
	\caption{Quantitative results of the ablation experiments. BL: the baseline model, SAR: the spatial attention to modify the reference image method, DT: the proposed Vision Transformer and CNN combined module.}
	\setlength{\abovecaptionskip}{0cm}
	\setlength{\belowcaptionskip}{0cm}
	\centering
	\begin{tabular*}{\hsize}{c@{\extracolsep{\fill}}c@{\extracolsep{\fill}}c@{\extracolsep{\fill}}c@{\extracolsep{\fill}}c@{\extracolsep{\fill}}c@{\extracolsep{\fill}}}
		\hline
		BL	&SAR	&DT	&${\rm PSNR}_{\mu}$	&${\rm PSNR}_{l}$	&HDR-VDP-2 \\
		\hline
		\checkmark &  &   & 44.21 	&42.17	&66.03 \\
		\checkmark & \checkmark &  &  44.28 & 42.26&  66.01\\
		\checkmark &	 & \checkmark & 44.31& 42.53& 66.06\\
		\checkmark & \checkmark &\checkmark  & 44.36 &42.73 & 66.08\\
		\hline
	\end{tabular*}
	\label{table 3}
\end{table}

{\bf – Baseline.} We take the HDR-Transformer \cite{b22}, which is constituted with a spatial attention (SA) module in shallow features extraction and Context-aware Vision Transformer (CA-ViT) encoders, as our baseline model. The baseline model keeps the same settings for training and testing as our proposed HDT-HDR.

{\bf – + SAR.} In this variant, instead of using spatial attention only to align non-reference images, we add spatial attention to be applied to reference images (SAR).

{\bf – + DT.} In this variant, the CA-ViT encoder used in the baseline model is replaced by the proposed dual-branch combined Transformer with deformable CNNs module.

{\bf – + SAR + DT.} The entire network of the proposed HDT-HDR.

As shown in Table \ref{table 3}, when we use SAR or DT on BL, it will increase the performance of the model, and the gain of DT is more significant than SAR. When both SAR and DT are adopted, the performance of the model reaches the best.

\section{Conclusions}
In this paper, we demonstrate several HDR imaging results in different scenes. State-of-the-art methods do not perform well enough in small motion deblurring or large motion deghosting. We propose a dual-branch Transformer that combines Transformers and deformable CNNs to overcome the lack of global information in local feature extraction, while global features are mainly extracted in vanilla ViTs. We first rectify the reference images by applying the spatial attention in the shallow feature extraction to avoid missing some information of non-reference images that could be used to recover the oversaturated/undersaturated region. Furthermore, we extend the channel attention by adding several deformable convolutional layers in the local branch, so that the feature extraction would not be limited to a certain kernel size, making it capable of capturing small movements in the local. Finally, we present HDT-HDR, which combines the advantages of Transformer and CNNs. Thus, our HDT-HDR has both the ability to remove ghosts caused by long-range movements and to remove blur caused by small motion. Extensive experiments show that the proposed method is quantitatively and qualitatively superior to state-of-the-art methods.

\end{document}